%% file: main.tex
\pdfoutput=1

\documentclass[11pt]{article}

\usepackage[preprint]{acl}

\usepackage{times}
\usepackage{latexsym}

\usepackage[T1]{fontenc}

\usepackage[utf8]{inputenc}

\usepackage{microtype}

\usepackage{inconsolata}

\usepackage{graphicx}

\usepackage{tcolorbox}
\usepackage{amsfonts} 
\usepackage{booktabs}
\usepackage{algorithm}
\usepackage{makecell}
\usepackage{subfigure}
\usepackage{subcaption}
\usepackage{algorithmic}
\usepackage{amsmath}
\usepackage{multirow}
\usepackage{colortbl}
\usepackage{xcolor}
\usepackage{arydshln}
\usepackage{pdflscape}
\usepackage{rotating}

\title{Generate, Not Recommend: Personalized Multimodal Content Generation }

\author{Jiongnan Liu$^{1,2}$, Zhicheng Dou$^{2}$, Ning Hu$^{3}$, Chenyan Xiong$^{1,*}$  \\
$^1$School of Computer Science, Carnegie Mellon University \\
$^2$Gaoling School of Artificial Intelligence, Renmin University of China \\
$^3$Serendipity One Inc. \\
\texttt{\{jiongnal, cx\}@andrew.cmu.edu}
}

\begin{document}
\maketitle
\def\thefootnote{*}\footnotetext{Corresponding author.}

\input{Abstract}

\input{Introduction}

\input{Method}

\input{Experimental_Methodology}

\input{Evaluation_Results}

\input{Conclusion}

\input{Limitations}

\bibliography{custom}

\input{Appendix}

\end{document}

%% file: Abstract.tex
\begin{abstract}
To address the challenge of information overload from massive web contents, recommender systems are widely applied to retrieve and present personalized results for users.
However, recommendation tasks are inherently constrained to filtering existing items and lack the ability to generate novel concepts, limiting their capacity to fully satisfy user demands and preferences.
In this paper, we propose a new paradigm that goes beyond content filtering and selecting: directly generating personalized items in a multimodal form, such as images, tailored to individual users.
To accomplish this, we leverage any-to-any Large Multimodal Models (LMMs) and train them in both supervised fine-tuning and online reinforcement learning strategy to equip them with the ability to yield tailored next items for users.
Experiments on two benchmark datasets and user study confirm the efficacy of the proposed method. Notably, the generated images not only align well with users’ historical preferences but also exhibit relevance to their potential future interests.

\end{abstract}

%% file: Introduction.tex
\section{Introduction}

\begin{figure}[t]
    \centering
    \includegraphics[width=0.95\columnwidth]{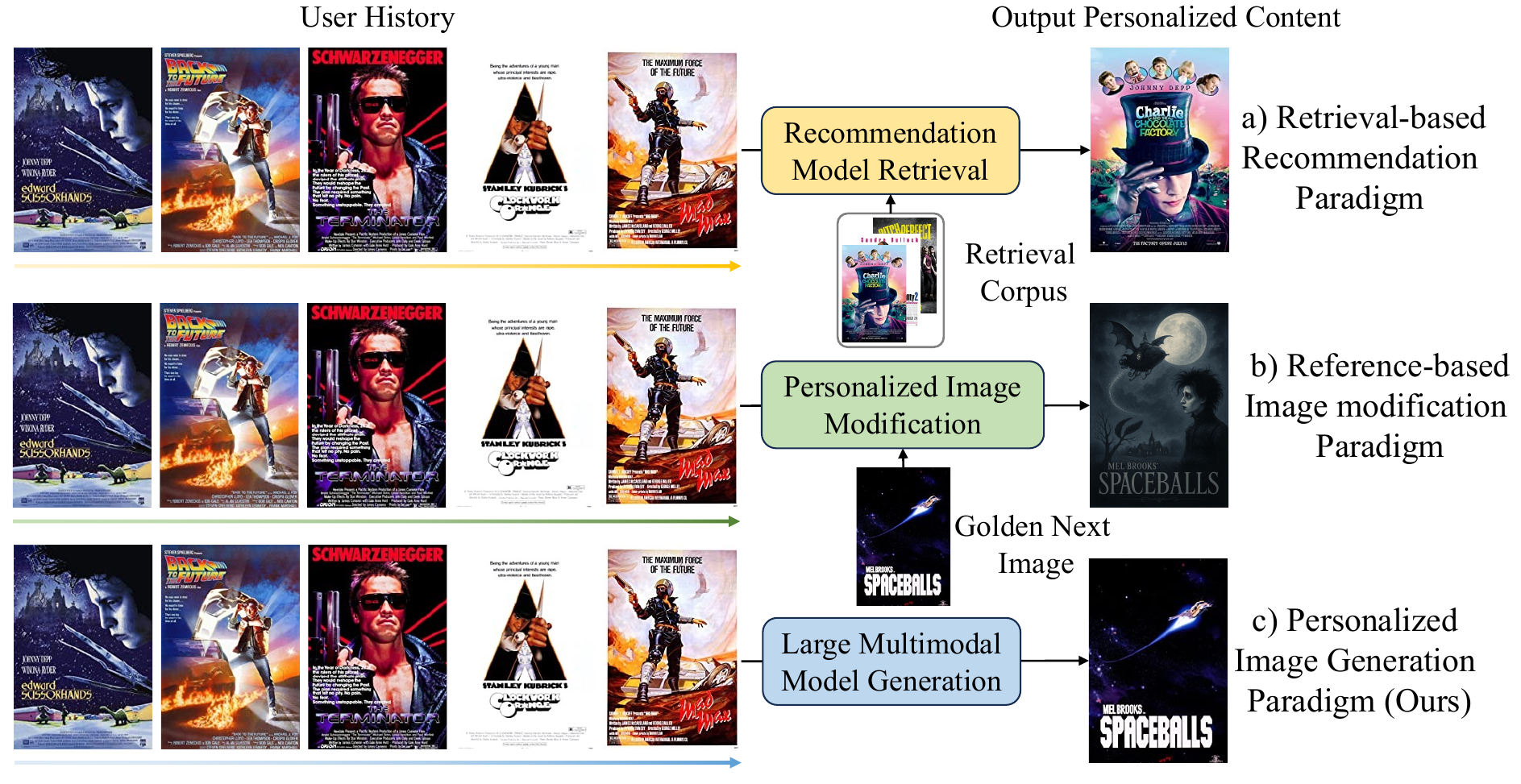}
    \caption{Personalized content provision paradigms.}
    \label{fig:intro}
\end{figure}

With the explosive growth of online content and documents, users are increasingly overwhelmed by the large amount of information available across platforms, leading to serious information overload~\cite{overload}. To alleviate it, recommender systems~\cite{sasrec,hllm,TASTE} have attracted huge attention from both academia and industry for selecting and displaying content aligned with users' interests and preferences based on their profiles and historical interactions. While effective in many scenarios, this retrieval-based paradigm may fails to fully fulfill user needs since they are limited to utilizing the existing contents, as shown in Figure~\ref{fig:intro}~a). To push beyond these limitations, a promising direction is to proactively generate content that users are likely to prefer next, particularly in multimodal formats such as images and videos, which are more attractive to users.

\label{sec:intro}

Several prior studies~\cite{pig,pmg,drc,iamg} have attempted to address similar problems by leveraging LMMs to summary user histories into profiles to guide downstream image generation modules to do personalized generation. However, they typically rely on the existing next candidate item as references and tailor them into a modified version to enhance personalization, as shown in Figure~\ref{fig:intro}~b).  Unfortunately, the golden following content that a user will interact with is often unavailable in practice, making these reference-dependent approaches inapplicable. In contrast, in this paper, we focus on a more realistic circumstance: generating personalized images solely based on users’ historical interactions without relying on any candidate or reference content, as shown in Figure~\ref{fig:intro}~c).

Achieving this requires the ability to understand user preferences in multimodal histories and to generate high-quality personalized content accordingly. To this end, our approach is built on any-modality-to-any-modality (any-to-any for short) LMMs, which we adapt to the end-to-end personalized generation setting without downstream image generation models through a two-stage training framework. First, relying on the user history data in recommendation benchmarks, we apply supervised fine-tuning, training the LMM to generate the ground-truth next-interacted items of users based on their histories. Second, we introduce the Group Relative Policy Optimization (GRPO) algorithm~\cite{grpo} as the online reinforcement learning stage to further enhance the generation quality. Specifically, we incorporate the personal relevance of generated images to the historical interacted, the actual next, and the potential future contents as GRPO's reward signals.

To evaluate the proposed image generation paradigm, we resort to two recommendation datasets in different domains and develop several semantic and perceptual similarity metrics to assess the personal relevance of the created content, together with the aesthetics metric. Experiments on these two benchmarks demonstrate that our approach consistently generates images that are more personally relevant both to users’ past and potential future preferences, than content retrieved by state-of-the-art recommendation models.

%% file: Method.tex
\section{Methodology}

To better meet user personalized needs, we propose a generative personalization paradigm that proactively creates tailored multimodal-form content for users based on their histories.
Specifically, we leverage any-to-any LMMs to encode user histories with both text and image content and generate images in an end-to-end fashion, which can be more effective than previous profile summarization with a conditional image generation pipeline.
To achieve this, we first utilize recommendation datasets to perform supervised fine-tuning on LMMs to generate the images of the users' ground-truth next items using their multimodal histories. 
Furthermore, to ease the lack of training data and enhance personalizations abilities, we apply the GRPO algorithm to conduct online reinforcement learning.

\subsection{Supervised Fine-tuning Step}
In our training pipeline, the supervised fine-tuning (SFT) stage equips LMMs with the basic ability to understand multimodal user histories and generate personalized images accordingly. Without existing dataset for assessing personalized image generation tasks, we utilize the user history data $\mathcal{H}_u=[\mathcal{H}_{u_1},\cdots,\mathcal{H}_{u_N}]$ from standard recommendation datasets, where each history item $\mathcal{H}_{u_i}$ consists of both text attributes $\mathcal{T}_{u_i}$ (\textit{e.g.}, movie titles) and image information $\mathcal{I}_{u_i}$ (\textit{e.g.}, movie posters) to cook our fine-tuning data. 
Specifically, the model is trained with the target of generating the visual tokens of image $\mathcal{I}_{u_k}$ of the $k$-th item using the multimodal contents of the previous $k-1$ items:
\begin{equation}
\small
\begin{aligned}
    \mathcal{L}_{\mathrm{SFT}} = -\!\!\sum_{u,k,j}\!  \log p_{\textrm{LMM}}(\mathcal{I}_{{u_{k}};j}|\mathcal{P}(\mathcal{H}_u[1:k-1]);I_{{u_{k}};<j}), \nonumber
\end{aligned}
\end{equation}
where $p_{\textrm{LMM}}$ denotes the predicted token distributions output by the LMM, $\mathcal{I}_{{u_{k}};j}$ denotes the $j$-th image token of the image of $k$-th history item, $\mathcal{P}$ denotes the instruction template for input construction (The detail template can be found in Appendix~\ref{apd:temp}). To mitigate the sparsity nature of recommendation datasets, we enrich the training data by randomly masking and swapping items in user histories.

\subsection{Online Reinforcement Learning Step}
\label{subsec:grpo}
It is inadequate for LMMs to learn the personalized generation task solely with the limited user-item interaction data.  For instance, an item that is strongly relevant to a user's histories and fits her interests may be absent from the corresponding interaction history due to a lack of exposure or logging omissions. To address this limitation, we introduce an online reinforcement learning (RL) stage that enables model to explore and optimize personalized generation beyond the observed data.

Specifically, we apply the GRPO algorithm~\cite{grpo}, where the rewards are calculated based on the aesthetics quality and the personal relevance of generated images. These relevance signals include similarities to a user’s past interactions, actual next items, and even potential future interests. The GRPO objective is defined as:
\begin{equation}
\label{eq:grpo}
\small
\begin{aligned}
    \mathcal{L}_{\mathrm{GRPO}} &= \mathbb{E}_{[\mathcal{H}_{u[:k]}\sim P(\mathcal{H}), \{\mathcal{I}_j\}_{j=1}^{G} \sim \pi_{\theta_{old}} (\mathcal{I} | \mathcal{H})]} \\
                                \frac{1}{G} \sum_{j=1}^{G} & \left[\min \left( \omega_j A_j, {\textrm{clip}}\left(\omega_j , 1\pm\epsilon\right) A_j \right) - \beta \mathbb{D}_{\mathrm{KL}} \left(\pi_\theta || \pi_{\textrm{ref}}\right) \right], \\
                                \omega_j & = \frac{\pi_{\theta} (\mathcal{I}_j | \mathcal{H})}{\pi_{\theta_{old}} (\mathcal{I}_j | \mathcal{H})}, \; 
                                A_j = \frac{R_j - {\textrm{mean}}(\{R_1,R_2,\cdots,R_G\})}{\textrm{std}(\{R_1,R_2,\cdots,R_G\})}, \nonumber
\end{aligned} 
\end{equation}
where $\pi_\theta$ is the current LMM policy, $\pi_{\theta_{old}}$ is the previous policy used for sampling, and $\pi_{\textrm{ref}}$ is a reference policy for KL regularization. The reward $r_j$ for each sampled image reflects a composite of personalization and quality scores. Further details on reward computation are described in Section~\ref{subsec:metric}.

%% file: Experimental_Methodology.tex
\section{Experimental Setup}

We introduce our experimental setup in this section.

\paragraph{Datasets \& Metrics} Given the lack of personalized generation benchmarks, we experiment on two existing recommendation datasets, MovieLens~\cite{movielens} and PixelRec~\cite{pixelrec}, which collect user movie and micro-video viewing records, respectively. We formulate the task as generating corresponding visual content: movie posters and micro-video covers. Dataset details are provided in Appendix~\ref{apd:data}. 

\label{subsec:metric}
Following reference-based personalized multimodal generation works~\cite{pig,pmg}, we measure the {personalized relevance} of the generated images by computing their semantic and perceptual similarity to the history, the golden next, and the potential future content. Specifically, Given the user history $\mathcal{H}_u=[\mathcal{H}_{u_1},\cdots,\mathcal{H}_{u_{k-1}}]$, we compare our generated image $\hat{\mathcal{I}_{u_{k}}}$ to {historical} item ${\mathcal{H}_{u_{1:k-1}}}$, the {golden} next item ${\mathcal{H}_{u_{k}}}$, and the potential future items  ${\mathcal{H}_{u_{k+1:k+p}}}$\footnote{In our evaluation, we set $k=6,p=3$.}.
Besides the personal relevance metrics, we also include an image aesthetics metric. 

We include the following metrics: \textbf{CTS/CIS} metrics stand for the CLIP~\cite{clip} matching score, computing the similarities between the generated images to the {text} or {image} contents of the compared items. \textbf{DIS} metric applies the DINO~\cite{dino} to calculate the image similarities between the generated ones and the compared ones. \textbf{LPIPS/MS-SSIM/SSIM} metrics quantify the perceptual visual personalization score for generated images with compared images using LPIPS~\cite{lpips}, MS-SSIM and SSIM~\cite{msssim}. Particularly, for the personalized relevance to historical items, we design an additional \textbf{PCS} metric, which calculates the clip scores between our generated images and the user profiles provided by the \href{https://huggingface.co/Qwen/Qwen2.5-VL-7B-Instruct}{Qwen-2.5-VL 7B} model based on user histories. We use \textbf{NIMA}~\cite{nima} model to evaluate the aesthetics of the generated images.

All the above metrics also serve as reward signals in our GRPO algorithm (Section~\ref{subsec:grpo}).

\paragraph{Baselines}
We compare our model against a multimodal adaptation of the state-of-the-art hierarchical LLM-based recommendation model HLLM~\cite{hllm}, in which \href{https://huggingface.co/Qwen/Qwen2-VL-2B-Instruct}{Qwen2-VL-2B}~\cite{qwen2vl} is used as both the user and item LLM. For our method, we leverage the \href{https://huggingface.co/deepseek-ai/Janus-Pro-1B}{Janus-Pro-1B}~\cite{janus} as our backbone LMM. As we show in Appendix~\ref{apd:vanilla}, the vanilla Janus model fails to generate semantically meaningful images, so we don't include it as our baseline. For both of these methods, we report the best metrics in the top 4 recommended/generated images. Implementation details are included in Appendix~\ref{apd:imp}.


%% file: Evaluation_Results.tex
\section{Evaluation Results}
We show experiments and analyses in this section.

\subsection{Overall Performance}
\input{Tables/overall_result}
\label{sec:result:overall}

Table~\ref{tlb:overall} reports the best scores achieved among the top-4 generated or retrieved images. Generally, after the SFT and online RL training stages, both the personal relevance and aesthetic quality of our generated images outperform the recommendation-retrieved ones. Furthermore, we observe that:

1) Our model achieves stronger results on the MovieLens dataset compared to PixelRec. This may be attributed to the fact that pretrained LMMs are more familiar with popular movie content, which may be included in their pretraining data. In contrast, LMMs tend not to be familiar with the micro-video contents. 

2) The GRPO-based online RL stage significantly boosts performance on metrics where the SFT-only model lags behind, such as the CIS score, demonstrating that enabling LMMs to explore and adapt through RL is effective for personalized generation tasks.

3) Our model not only achieves improvements on personalization metrics related to historical items but also on those related to potential future items. This suggests that our model can capture long-term user interests effectively.

\subsection{Further Analysis}
\noindent\textbf{User study}\quad
To further evaluate the effectiveness of the online RL training stage, we conducted a user study with three independent annotators. They were asked to assess 30 generation pairs from the MovieLens dataset, produced by the Janus+SFT and Janus+SFT+RL models. For each case, we presented all 8 posters (4 from each model) in a randomized order and asked the annotators to label the posters they considered to be good generations. The final preference for each case was determined by majority vote based on which model produced more high-quality posters. As shown in Table~\ref{table:user_study}, the GRPO-based RL training significantly improves the quality of personalized image generation.

\begin{table}[t]
\caption{Pairwise user study on MovieLens dataset.}
\centering\resizebox{1.0\linewidth}{!}{
\begin{tabular}{lcccc}
\toprule
Model & Win     & Lose    & Tie     & Win+Tie \\
\midrule 
Jansu+SFT  & 36.67\% & 53.33\%  & 10\% & 46.67\%\\
Janus+SFT+RL & \textbf{53.33\%} & 36.67\% & 10\% & \textbf{63.33\%} \\

\bottomrule
\end{tabular}}\label{table:user_study}
\end{table}


\begin{figure}[t]
    \centering
    \includegraphics[width=0.99\columnwidth]{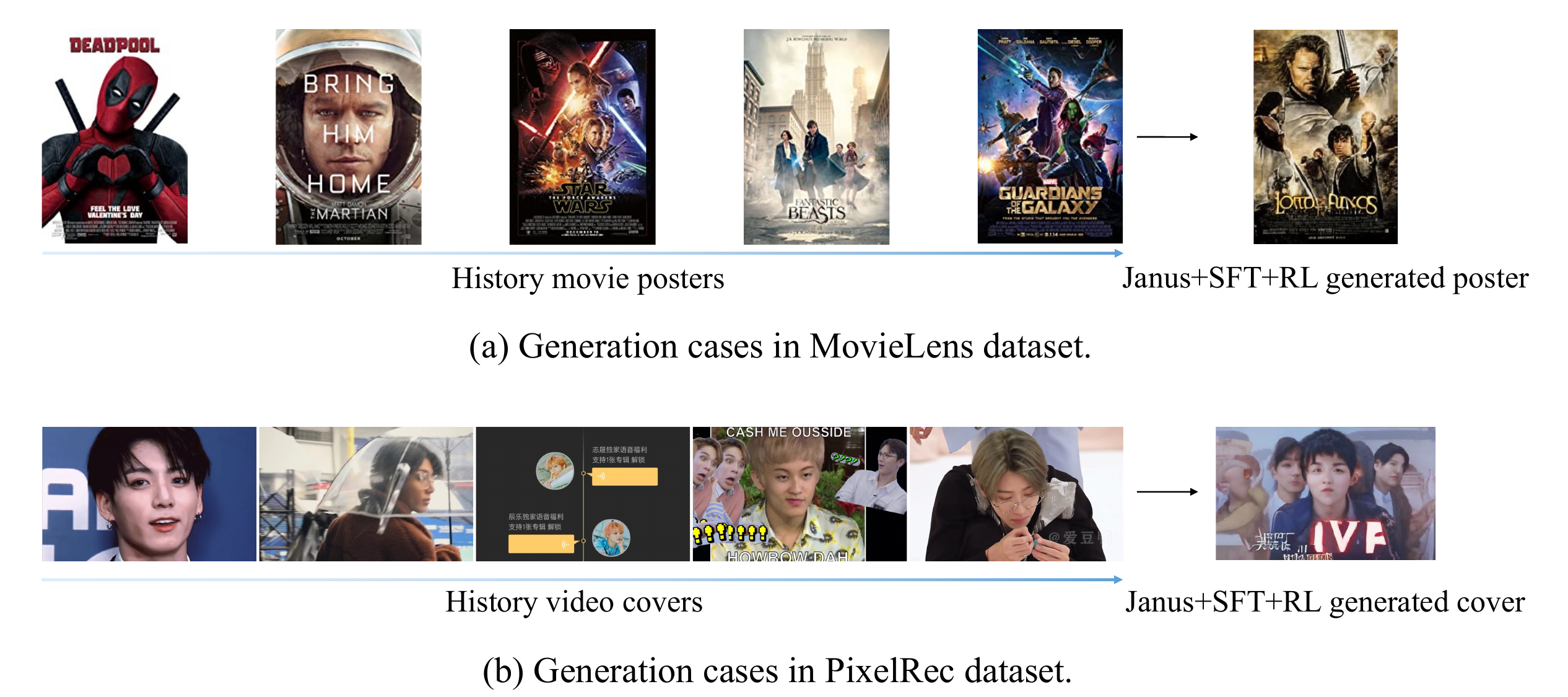}
    \caption{Generation cases in both datasets.}
    \label{fig:case}\vspace{-8pt}
\end{figure}

\noindent\textbf{Case study}\quad
In this section, we present several generation cases in Figure~\ref{fig:case}. The results demonstrate that our proposed method can effectively generate personalized posters or video covers based on users’ historical preferences.
On the MovieLens dataset, for example, the user shows a clear interest in epic fantasy and science fiction franchises. Accordingly, our method generates a poster for The Lord of the Rings, aligning well with the user’s taste.
On the Pixelrec dataset, the user appears to prefer videos featuring idols, particularly those with close-up facial covers. Interestingly, our method captures this preference and produces a video cover prominently featuring an idol’s face.

\begin{figure}[t]
    \centering
    \includegraphics[width=0.95\columnwidth]{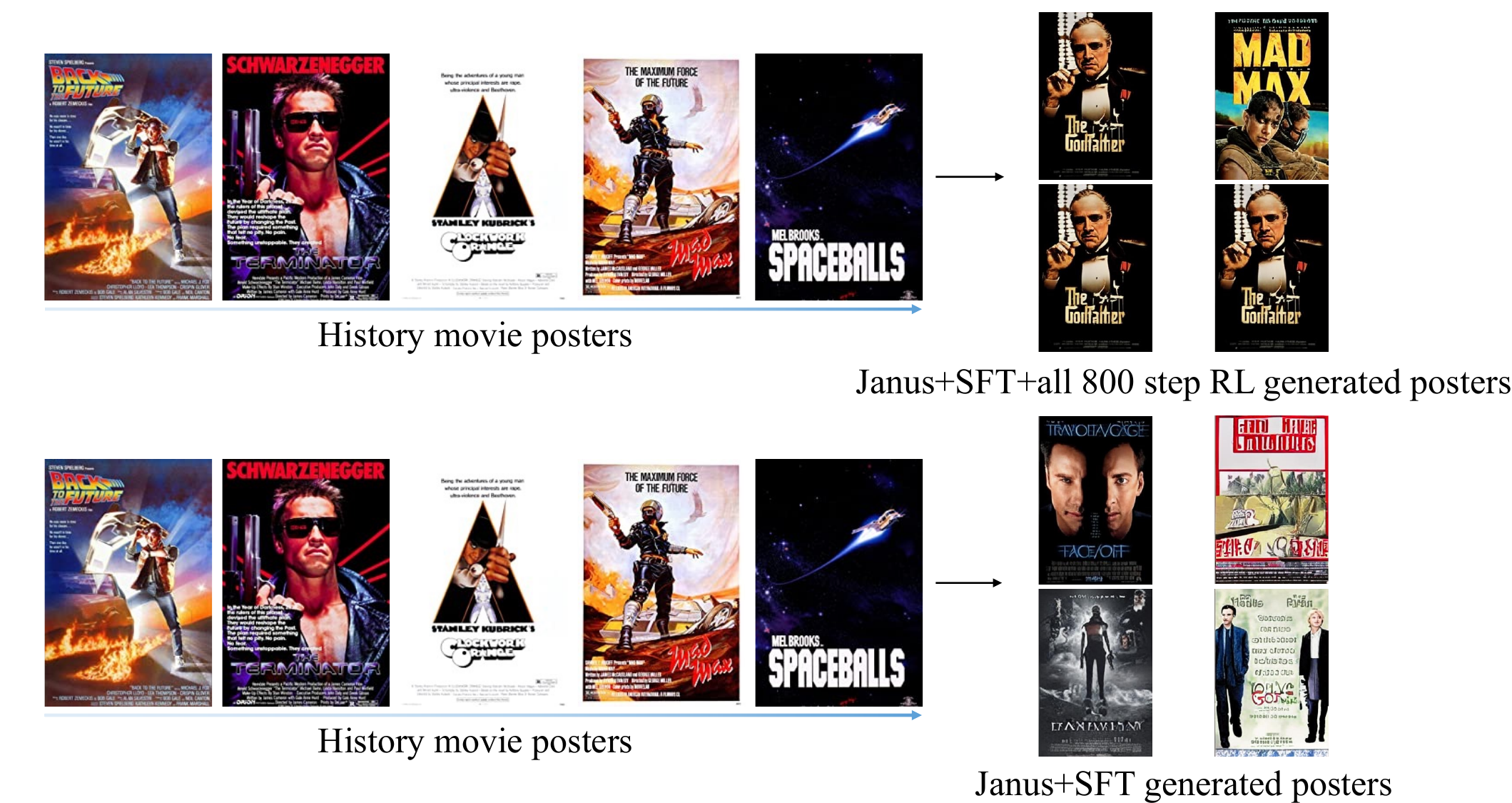}
    \caption{Generation cases in MovieLens dataset.}
    \label{fig:fail}
\end{figure}

\noindent\textbf{Reward Hacking Analysis}\quad
Figure~\ref{fig:fail} shows cases from models after all RL steps and after SFT stage. 
We find that the RL model tends to generate popular posters like The Godfather, a behavior not seen after SFT. 
We attribute this to the reward hacking phenomenon, where the relevance metrics favor popular images. For example, the Godfather poster has a higher average \textbf{CIS} reward, likely because the CLIP model was exposed to such popular images during pretraining. The model exploits this shortcut to maximize rewards, resulting in a popularity bias. This issue does not appear in the PixelRec dataset, as CLIP hasn't seen micro-video covers.

\begin{figure}[t]
    \centering
    \includegraphics[width=0.8\columnwidth]{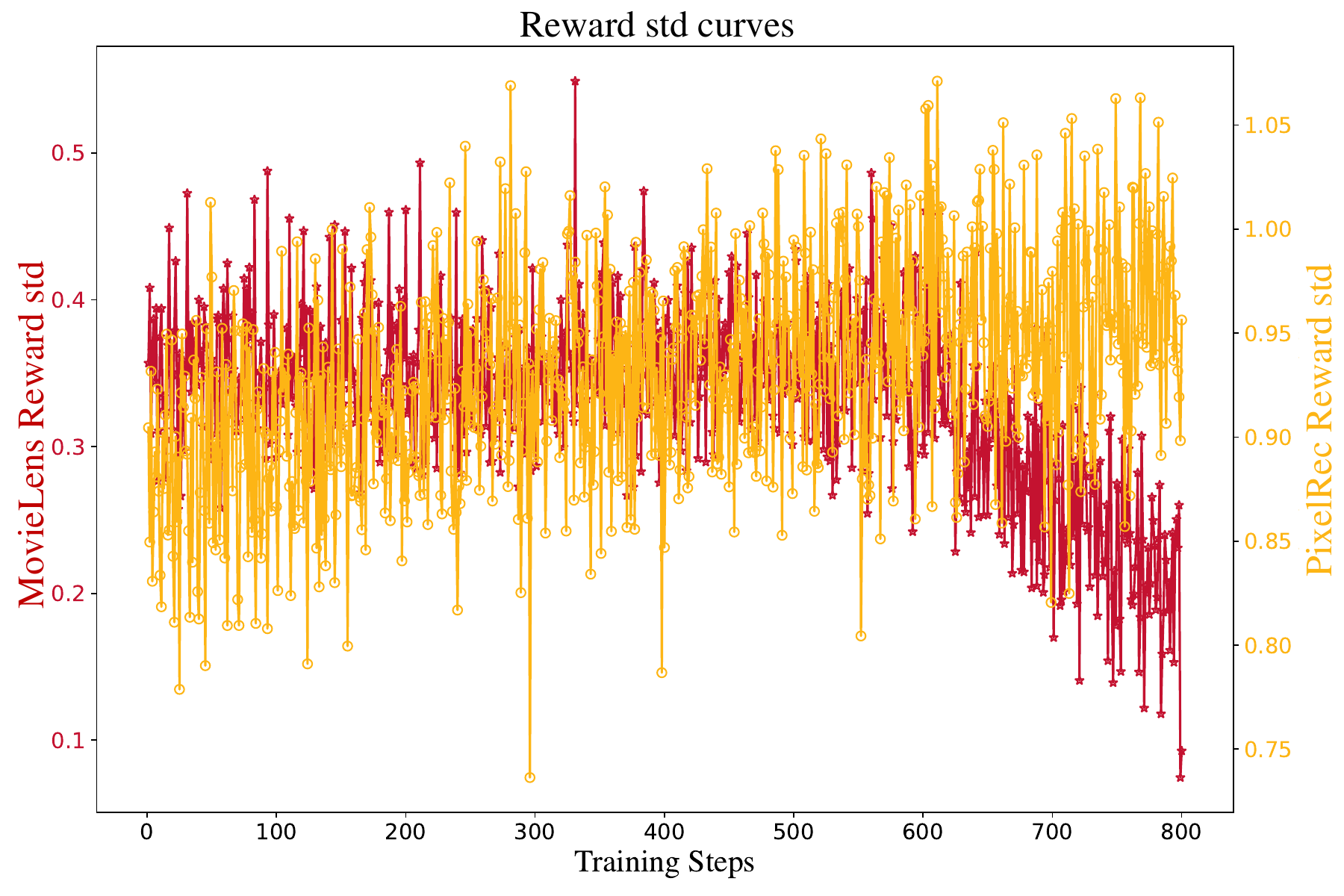}
    \caption{Reward standard deviation curves in datasets.}
    \label{fig:std_curve}\vspace{-8pt}
\end{figure}

The reward standard deviation  curves in Figure~\ref{fig:std_curve} further support this. In MovieLens, reward standard deviation drops suddenly during RL, indicating the model has discovered the shortcut and is repeatedly generating a few or even a single popular item. To avoid this, we select our final model before this. In contrast, the reward standard deviation remains stable in the PixelRec dataset. However, we need to state that reward hacking is an open problem in online RL with model-based reward, which we will explore in our future work.

%% file: Tables/overall_result.tex
\begin{table}[t]
 \centering
 \caption{The personalized generation/recommendation performance on both datasets. The best results are in \textbf{bold} while the second best results are \underline{underlined}.} \label{tlb:overall}
   \setlength{\tabcolsep}{1.1mm}\resizebox{1.0\linewidth}{!}{
  \begin{tabular}{clcccccc}
  	\toprule
        \multicolumn{2}{c}{\multirow{3}{*}{{Models}}} & \multicolumn{3}{c}{MovieLens} & \multicolumn{3}{c}{PixelRec} \\
        \cmidrule(lr){3-5} \cmidrule(lr){6-8} 
        & &  \makecell[c]{HLLM} & \makecell[c]{Janus\\+SFT} & \makecell[c]{Janus\\+SFT+RL}&  \makecell[c]{HLLM} & \makecell[c]{Janus\\+SFT} & \makecell[c]{Janus\\+SFT+RL} \\
        \midrule
        \multirow{3}{*}{\makecell[c]{Relevance to\\ Golden Next \\ Item}}
        & CTS$\uparrow$ & \underline{19.28} & \textbf{20.00} & {18.78}   & \underline{21.04} & {20.54} & \textbf{22.28} \\
        & CIS$\uparrow$ & \underline{47.32} & {40.00}  & \textbf{50.82} & \textbf{39.95} & {34.67}  & \underline{39.63} \\
        & DIS$\uparrow$ & \textbf{32.95} & {28.85}  &  \underline{30.26} & \textbf{22.01} & {21.04}  & \underline{21.35} \\
        \midrule
        \multirow{7}{*}{\makecell[c]{Relevance to\\ Historical \\ Items}}
        & CTS$\uparrow$  & \underline{18.38} & \textbf{18.56} & {17.76}  & \underline{19.95} & {19.24} & \textbf{21.36}  \\
        & CIS$\uparrow$  & \underline{46.87} & {39.21} & \textbf{50.29}  & \textbf{38.92} & {33.04} & \underline{38.41}  \\
        & DIS$\uparrow$  & \textbf{31.16} & {27.06} & \underline{28.53}  & \textbf{20.57} & {18.43} & \underline{18.80} \\
        & PCS$\uparrow$  & \textbf{22.98} & {21.31} & \underline{21.93}  & \textbf{21.04} & {19.61} & \underline{20.81} \\
        & LPIPS$^{*}\downarrow$ & \underline{73.58} & {73.81} & \textbf{72.82} & \underline{75.53} & \textbf{75.45} & {75.96} \\
        & SSIM$^{*}\uparrow$  & \underline{23.91} & \textbf{24.95} & {22.89}  & \underline{35.61} & {35.23} & \textbf{36.64} \\
        & MS-SSIM$^{*}\uparrow$  &  {11.46} & \textbf{12.22} & \underline{11.70}  & \underline{14.99} & {14.81} & \textbf{15.89} \\
        \midrule
        \multirow{6}{*}{\makecell[c]{Relevance to\\ Potential Future \\ Items}} 
        & CTS$\uparrow$  & {17.75}  & \textbf{18.78} & \underline{17.88} & \underline{19.88} & {19.42} & \textbf{21.38} \\
        & CIS$\uparrow$  & \underline{45.68} & {38.74} & \textbf{49.60} & \underline{37.95} & {33.09} & \textbf{38.40} \\
        & DIS$\uparrow$  & \textbf{30.43} & {27.15} & \underline{28.61}  & \textbf{19.73} & {19.28} & \underline{19.41} \\
        & LPIPS$^{*}\downarrow$  & \underline{73.63} & {73.86} & \textbf{73.13}  & \underline{75.34} & \textbf{75.16} & {75.61}\\
        & SSIM$^{*}\uparrow$ & \underline{24.00} & \textbf{25.22} & {23.23} & \underline{35.59} & {35.32} & \textbf{36.72} \\
        & MS-SSIM$^{*}\uparrow$ & {12.11} & \textbf{13.03} & \underline{12.38} & \underline{15.86} & {15.61} & \textbf{16.76} \\
        \midrule
        Aesthetics Metric & NIMA$\uparrow$ &  {5.582} & \underline{5.783} & \textbf{6.039} & \textbf{6.086} & \underline{6.085} & {5.914} \\
        \midrule
        \multicolumn{2}{c}{Metric Average${^{\dagger}}$}  & \underline{26.04} & 24.31 & \textbf{26.16} & \underline{24.61} & 23.22 & \textbf{24.83}\\
        \bottomrule
  \end{tabular}}
    \raggedright
    \footnotesize{$^{*}$ For better comparison, we multiply these metrics by 100.}
    
    \footnotesize{$^{\dagger}$ We use 100 to subtract the LPIPS metric in metric average.}
\end{table}

%% file: Conclusion.tex
\section{Conclusion}

This paper explores a new paradigm for personalized multimodal content generation without relying on any existing reference. 
Built on any-to-any LMMs such as Janus-Pro, our model generally yield better results compared with the state-of-the-art recommendation models falling in the retrieval paradigm.
As a preliminary exploration into personalized generalization, this work opens up new possibilities for user-centric content creation. We hope it can serve as an inspiration for future research in this direction.

%% file: Limitations.tex
\section*{Limitations}

In this paper, we explore the personalized multimodal content generation tasks to revolutionize the current recommendation-based personalized content retrieval. Experimental results confirm the potential applications of this novel paradigm. Despite the promising results, our work faces several limitations for future exploration.

First, there is currently a lack of well-established benchmarks for personalized multimodal content generation tasks. As a result, we rely on existing recommendation datasets as a proxy, which may not fully capture the complexity of the task. Building dedicated benchmarks or even interactive evaluation platforms as arena remains a crucial step toward advancing this emerging research area.

Second, from the above case study and our failure analysis, we observe that the fine-tuned Janus model tends to generate images that closely resemble those seen during training, indicating a limited ability to generalize beyond memorized visual patterns. This suggests that the current backbone LMMs struggle to understand multimodal inputs and generate multimodal outputs simultaneously, which is also shown in Appendix~\ref{apd:vanilla}. For example, the Janus-Pro-1B model designs disentangle image understanding and generation modules and were pretrainined on image understanding and generation data independently. In future work, we aim to explore more powerful vision-language models with stronger cross-modal reasoning and generation capabilities to better support personalized, open-ended image generation.

\section*{Ethical Statements}
In this paper, we introduce a framework for personalized multimodal content generation. Our experiments are conducted on widely-used public recommendation datasets (MovieLens and PixelRec), which either consist of anonymized user interactions or are collected with proper user consent under clear terms of use. No personally identifiable information (PII) is involved in our model training or evaluation. Based on generative models, our framework has the potential to produce misleading or inappropriate content.  We acknowledge this limitation and will explore further safeguard techniques such as output filtering and generation constraints in our future work.

%% file: Appendix.tex
\appendix

\section{Dataset Details}
\label{apd:data}
Due to limited computational resources, we sampled subsets of the MovieLens and PixelRec datasets instead of using the full version.  In these datasets, we first sort user-item interactions in chronological order and apply sliding window of six interactions to generate benchmark data. For each data sample, we regard the first five interactions as user histories and the last one as the ground-truth times. We split the built samples into training, validation and test sets according to a ratio of 8:1:1. The statistics of the sampled datasets are shown in Table~\ref{tlb:sta}.

\begin{table}[!t]
\centering
\small
\caption{Dataset Statistics}\label{tlb:sta}
{\resizebox{0.99\linewidth}{!}{
\begin{tabular}{lrrr}
\toprule
Dataset & \#User & \#Item & \#Interaction \\
\midrule
MovieLens & 594 & 6,961 & 31,058 \\
PixelRec & 600 & 23,546 & 33,383 \\
\midrule
\end{tabular}}}
\end{table}

\section{Vanilla Janus Generation Cases}
\label{apd:vanilla}
\begin{figure}[t]
    \centering
    \includegraphics[width=0.95\columnwidth]{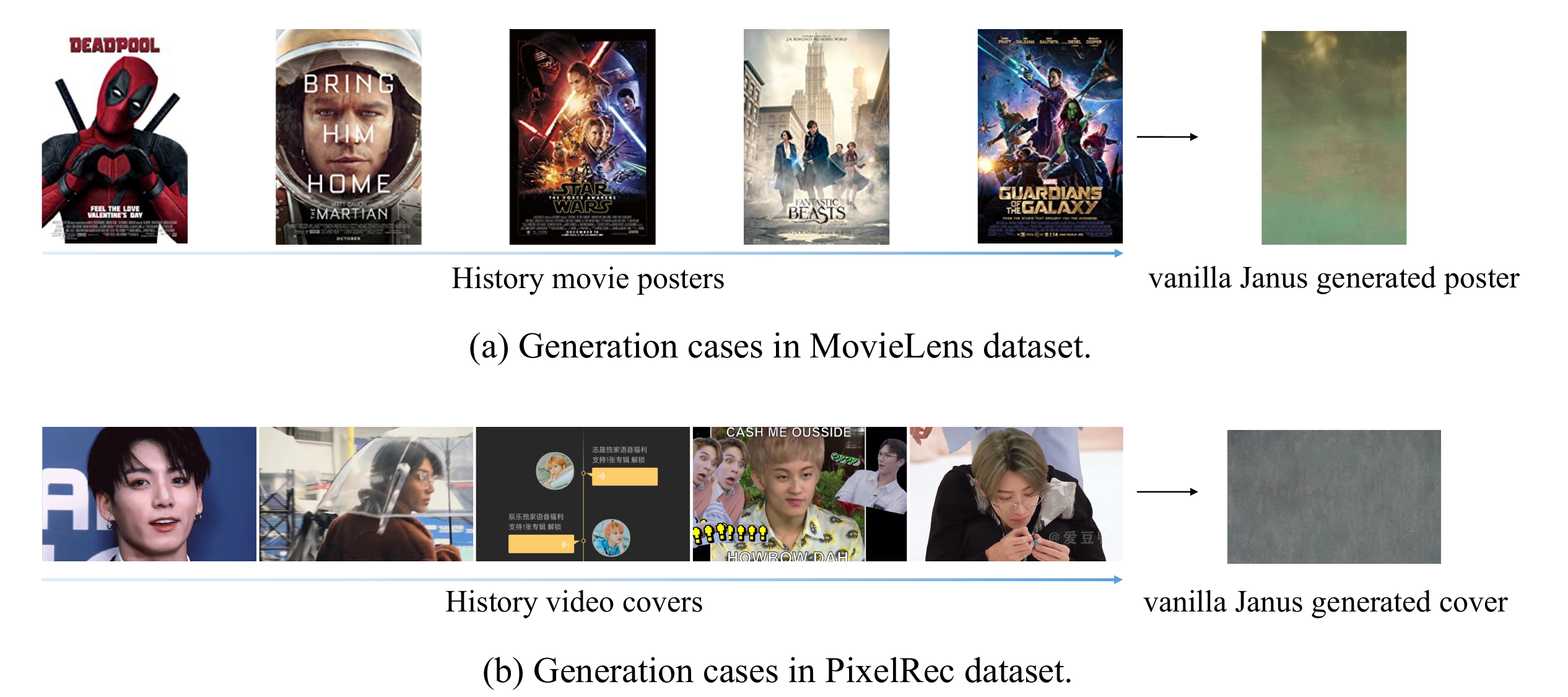}
    \caption{Generation cases by Vanilla Janus-Pro-1B model without further tuning.}
    \label{fig:vanilla}
\end{figure}

In this section, we show the generated images of the vanilla Janus-Pro-1B in Figure~\ref{fig:vanilla}. We can observe that the vanilla model without SFT and RL simply produces meaningless color blocks, which suggests that current models struggle in generating personalized multimodal contents.

\section{Implementaion Details}
\label{apd:imp}
Most LMMs, such as Qwen~\cite{qwen25vl,qwen2vl} and LLaVA~\cite{llava} can only perform image-to-text or text-to-image tasks. Since our task requires any-modality-to-any-modality ability, we leverage the Janus-Pro-1B~\cite{janus} as our backbone model. The training procedure consists of two stages:

\begin{itemize}
    \item \textbf{Supervised Fine-tuning (SFT)}: In the supervised fine-tuning stage, we train the model for one epoch with a learning rate of $4e^{-5}$ and a batch size of $64$. The best model is selected based on the lowest validation loss.
    \item \textbf{Online Reinforcement Learning}: We train the model for up to 800 steps using a learning rate of $1e^{-6}$ due to limited computational resources, rollout size $G=8$, and KL penalty coefficient $\beta=0.04$ in Equation in Section~\ref{subsec:grpo}. We use the model after the SFT stage as the reference policy $\pi_{\textrm{ref}}$. All reward metrics are scaled to the range $[0, 1]$ and added together. The RL batch size is set to 8. The final model is selected based on the highest average reward before any sign of reward hacking (e.g., sudden reward standard deviation drops).
\end{itemize}

For the HLLM baseline, we use the same training dataset as in our SFT stage with a learning rate of $1e^{-4}$. Noticed that for fair comparison, we use the VQ model in Janus-Pro-1B to encode and decode the recommended posters to ensure the same resolution and image quality.

\section{Instruction Template}
\label{apd:temp}
The instruction templates for two datasets are shown in Figure~\ref{box:movielens_template} and Figure~\ref{box:pixelrec_template}, respectively.
\begin{figure*}[h]
    \begin{tcolorbox}[title={Instruction template of Janus-Pro-1B to generate personalized movie posters on MovieLens dataset based on user histories.}] 
    Please generate a movie poster that would attract my interest. Here are the movies I have watched: \newline
    Movie 1: title: \{movie$_1$\ title\} genres: \{movie$_1$\ genres\} intro: \{movie$_1$ intro\} image: \{movie$_1$ poster\} \newline
    Movie 2: title: \{movie$_2$\ title\} genres: \{movie$_2$\ genres\} intro: \{movie$_2$ intro\} image: \{movie$_2$ poster\} \newline
    Movie 3: title: \{movie$_3$\ title\} genres: \{movie$_3$\ genres\} intro: \{movie$_3$ intro\} image: \{movie$_3$ poster\} \newline
    Movie 4: title: \{movie$_4$\ title\} genres: \{movie$_4$\ genres\} intro: \{movie$_4$ intro\} image: \{movie$_4$ poster\} \newline
    Movie 5: title: \{movie$_5$\ title\} genres: \{movie$_5$\ genres\} intro: \{movie$_5$ intro\} image: \{movie$_5$ poster\} \newline
    Design a new movie poster inspired by the text and visual elements of these films, making it appealing based on my viewing preferences.
    \end{tcolorbox}
\caption{Instruction template of Janus-Pro-1B to generate personalized movie posters on MovieLens dataset based on user histories}
\label{box:movielens_template}
\end{figure*}

\begin{figure*}[h]
    \begin{tcolorbox}[title={Instruction template of Janus-Pro-1B to generate personalized micro-video cover on PixelRec dataset based on user histories.}] 
    Please generate a video cover that would attract my interest. Here are the videos I have watched: \newline
    Video 1: title: \{video$_1$\ title\} tag: \{video$_1$\ tag\} description: \{video$_1$ description\} image: \{video$_1$ cover\} \newline
    Video 2: title: \{video$_2$\ title\} tag: \{video$_2$\ tag\} description: \{video$_2$ description\} image: \{video$_2$ cover\} \newline
    Video 3: title: \{video$_3$\ title\} tag: \{video$_3$\ tag\} description: \{video$_3$ description\} image: \{video$_3$ cover\} \newline
    Video 4: title: \{video$_4$\ title\} tag: \{video$_4$\ tag\} description: \{video$_4$ description\} image: \{video$_4$ cover\} \newline
    Video 5: title: \{video$_5$\ title\} tag: \{video$_5$\ tag\} description: \{video$_5$ description\} image: \{video$_5$ cover\} \newline
    Design a new video cover inspired by the text and visual elements of these videos, making it appealing based on my viewing preferences.
    \end{tcolorbox}
\caption{Instruction template of Janus-Pro-1B to generate personalized micro-video cover on PixelRec dataset based on user histories}
\label{box:pixelrec_template}
\end{figure*}

\section{Use of AI Assistants}

We use ChatGPT\footnote{\url{https://chatgpt.com/}} (powered by GPT-4o) to improve the presentations of this paper.